\documentclass[letterpaper, 10 pt, conference]{ieeeconf}

\usepackage{times}
\usepackage[pdftex]{graphicx}
\usepackage{amsmath,amssymb,amsopn,amstext,amsfonts}
\usepackage{cancel}
\usepackage[space]{cite}
\usepackage{pdfsync}
\usepackage{balance}
\usepackage{color}
\usepackage{mathtools}
\usepackage{algpseudocode}
\usepackage{algorithm}
\usepackage{bm}

\usepackage{diagbox}
\usepackage{float}
\usepackage{epstopdf}
\usepackage{pifont}
\usepackage{amsmath}
\usepackage{multirow}
\usepackage{url}
\usepackage{verbatim}
\usepackage{booktabs}
\usepackage{graphicx}
\usepackage{subcaption}
\usepackage{threeparttable}  
\usepackage{makecell}
\usepackage[linkcolor=black,citecolor=black,urlcolor=black,colorlinks=true]{hyperref}

\graphicspath{{figures/}}
\IEEEoverridecommandlockouts
\DeclareGraphicsExtensions{.pdf}

\title{\LARGE \bf EGO-Swarm: A Fully Autonomous and Decentralized Quadrotor Swarm System in Cluttered Environments}
\author{Xin~Zhou\textsuperscript{1,3}, Jiangchao~Zhu\textsuperscript{1,3}, Hongyu~Zhou\textsuperscript{3}, Chao~Xu\textsuperscript{1,2}, and~Fei~Gao\textsuperscript{1,2} 
	\thanks{ This work was supported by National Natural Science Foundation of China under Grant 62003299. }
	\thanks{\textsuperscript{1}State Key Laboratory of Industrial Control Technology, Institute of Cyber-Systems and Control, Zhejiang University, Hangzhou 310027, China.}
	\thanks{\textsuperscript{2}Huzhou Institute, Zhejiang University, Huzhou 313000, China.}
	\thanks{\textsuperscript{3}National Engineering Research Center for Industrial Automation (Ningbo Institute), Ningbo 315000, China.}
	\thanks{E-mail: {\tt\small $\{$iszhouxin, fgaoaa$\}$@zju.edu.cn}}
}

\begin{document}

\maketitle
\thispagestyle{empty}
\pagestyle{empty}

\begin{abstract}
This paper presents a decentralized and asynchronous systematic solution for multi-robot autonomous navigation in unknown obstacle-rich scenes using merely onboard resources. 
The planning system is formulated under gradient-based local planning framework, where collision avoidance is achieved by formulating the collision risk as a penalty of a nonlinear optimization problem.
In order to improve robustness and escape local minima, we incorporate a lightweight topological trajectory generation method.
Then agents generate safe, smooth, and dynamically feasible trajectories in only several milliseconds using an unreliable trajectory sharing network.
Relative localization drift among agents is corrected by using agent detection in depth images.
Our method is demonstrated in both simulation and real-world experiments.
The source code is released for the reference of the community.
\end{abstract}

\section{Introduction}
\label{sec:introduction}
The agility of quadrotors enables this machine to perform single-agent autonomous navigation in unknown environments and multi-agent precise formation control in open or known fields. 
However, few works combine both of them to present any real-world system which is capable of navigating quadrotor swarms sharing the same unknown space, especially with only onboard processing. 
The difficulties in deploying multiple quadrotors in unknown environments include but not limit to nontriviality of obstacle parameterization, limited sensing range, unreliable and bandwidth limited communication, and positioning drift caused by inconsistent localization. 
Several related works, e.g. \cite{mcguire2019minimal, zhou2018agile}, push the real-world quadrotor swarms capable of obstacle avoidance much further. 
However, above-mentioned difficulties are always neglected under motion capture systems or in pure simulation for most of the previous works, limiting the application of their algorithms in practice.

\begin{figure}[t]
	\centering
	\includegraphics[width=1.0\linewidth]{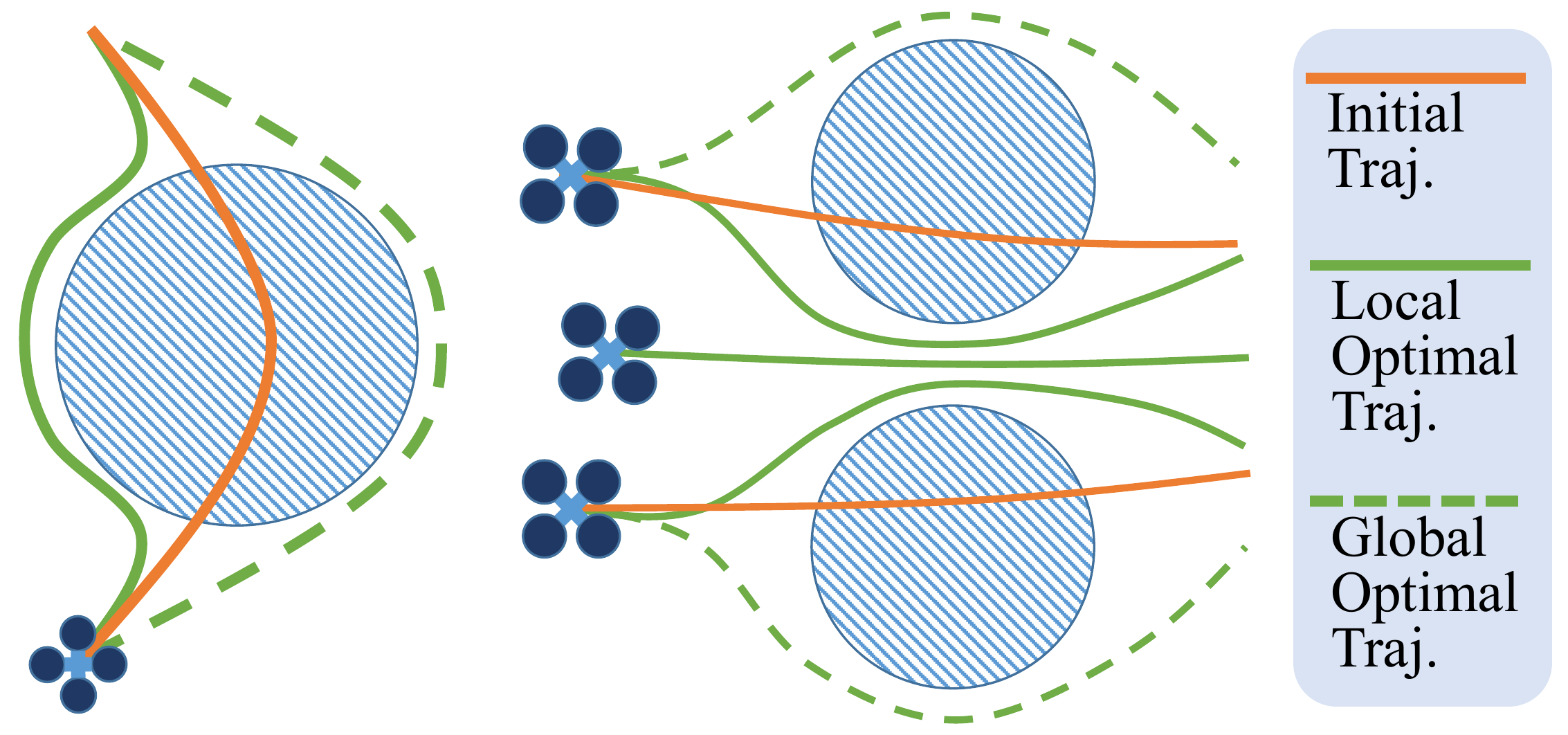}
	\captionsetup{font={small}}
	\caption{ \textbf{Left:} Dynamical infeasibility of the local optimal trajectory. \textbf{Right:} Three agents crowded through the flight corridor. }
	\label{pic:if_no_distinct}
	\vspace{-0.8cm}
\end{figure}

This paper proposes a systematic solution that enables high-performance traveling in cluttered environments for quadrotor swarms in the presence of these above-mentioned difficulties. 
In addition, it requires no external localization and computation or a pre-built map.
The proposed system named \textbf{EGO-Swarm} is an extension of our previous work \textbf{E}SDF-free \textbf{G}radient-based l\textbf{O}cal planner (EGO-Planner)\cite{zhou2020ego}, which lays a solid baseline for onboard local planning of a single quadrotor in unknown environments. 
The expansion consists of two parts, topological planning, and reciprocal collision avoidance.
As depicted in Fig.\ref{pic:if_no_distinct}, non-convex configuration space may lead to undesired behaviors, such as dynamical in-feasibility or crowded navigation.
Therefore, a strategy such as topological planning to escape local minima is beneficial. 
Based on the collision cost formulation in EGO-Planner, the front-end topological path search is done implicitly, therefore requiring nearly no computation. 
Decentralized reciprocal collision avoidance is achieved by adding a weighted penalty of swarm collision to the objective function.
This penalty is evaluated by comparing the agent distribution at a pried of future time with the being optimized trajectory.
%A stable sequence connection is used to synchronize timestamp and manage system startup.
In order to minimize data transmission and allow unreliable communication, a broadcast network is used to share trajectories.
To correct relative localization drift, which can increase to half a meter, observations of witnessed agents and predictions from trajectory evaluations are compared. 

\begin{figure}[t]
	\vspace{0.3cm}
	\centering
	\includegraphics[width=1.0\linewidth]{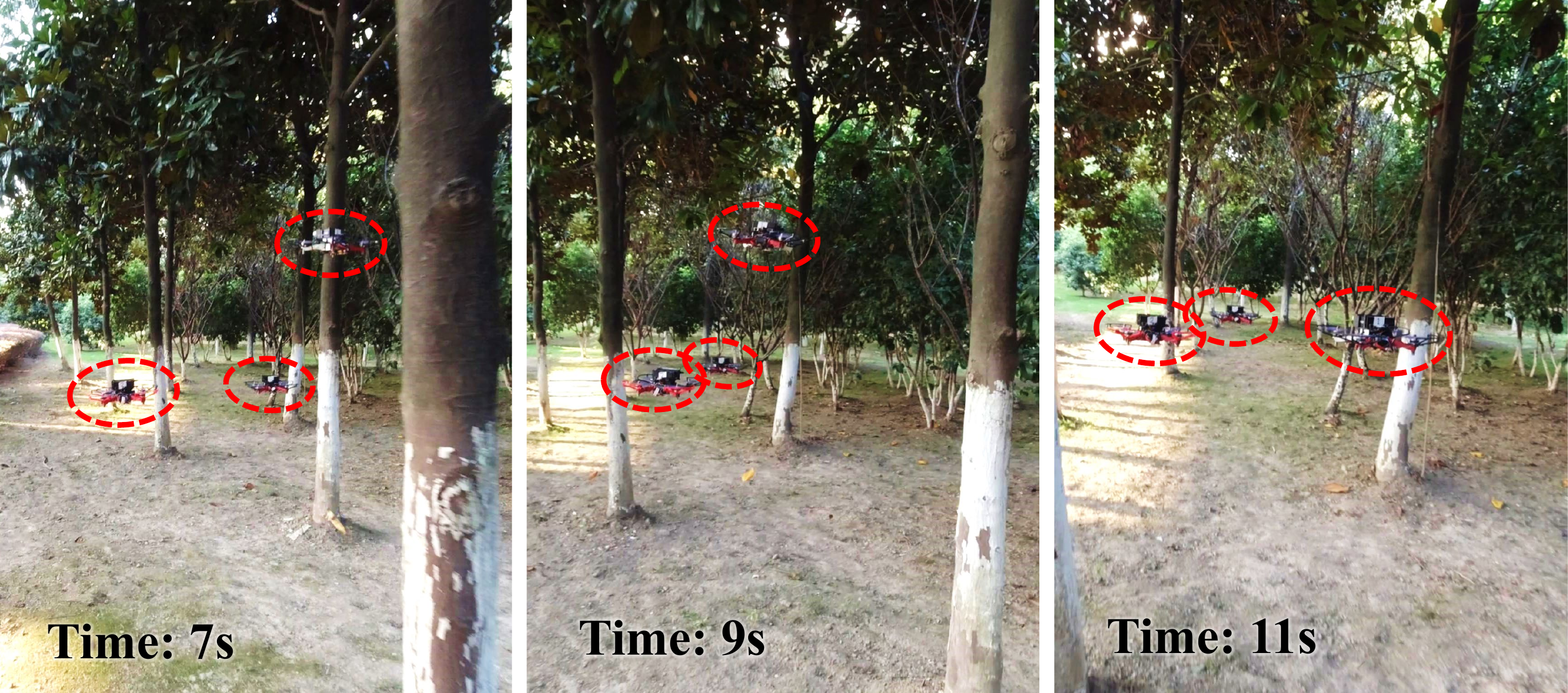}
	\captionsetup{font={small}}
	\caption{ Three snapshots of a flight in forest of three quadrotors. }
	\label{pic:forest}	
	\vspace{-0.3cm}
\end{figure}

\begin{figure}[t]
	\centering
	\includegraphics[width=1.0\linewidth]{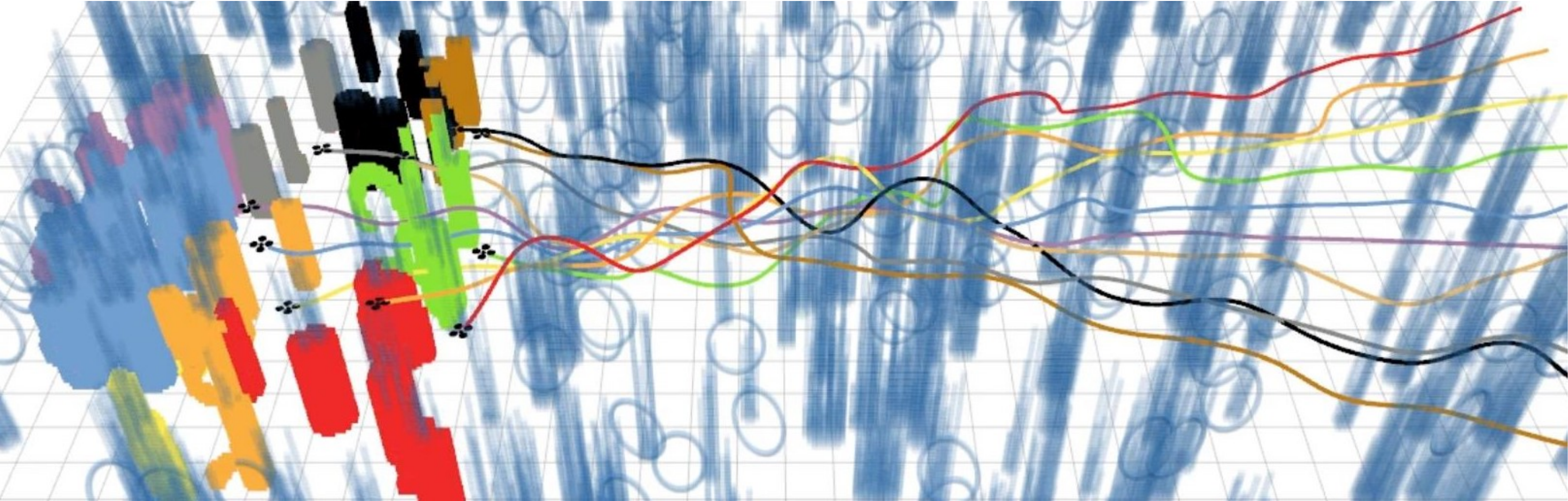}
	\captionsetup{font={small}}
	\caption{ Ten quadrotors fly in simulation. Colored blocks represent the local perception of quadrotors. Each color is associated with a quadrotor. }
	\label{pic:sim}	
	\vspace{-1.5cm}
\end{figure}

Real-world experiments are presented to validate our proposed swarm system.
To the best of our knowledge, this is the first systematic solution for fully autonomous decentralized quadrotor swarms in unknown cluttered environments, which means perception, planning, and control are integrated into the onboard system.
Comparisons with several SOTA methods show the computation efficiency and robustness.
The contributions of this paper are summarized as follows:

\begin{enumerate}
	\item 
	We extend our previous work EGO-Planner to propose a novel and robust topological planning method with nearly no extra computation.
	\item 
	We propose a decentralized and asynchronous quadrotor swarm framework which is insensitive to unreliable communication and localization drift.
	\item  
	The proposed method is integrated into a fully autonomous quadrotor system with hardware and software released for the reference of the community\footnote{https://github.com/ZJU-FAST-Lab/ego-planner-swarm}.  
\end{enumerate}

\section{Related Works}
\label{sec:related_works}
\subsection{Single Quadrotor Local Planning}

Gradient-based motion planning is the mainstream for quadrotor local planning.
Based on the pioneering works \cite{ratliff2009chomp, kalakrishnan2011stomp} that formulate the local planning problem as unconstrained nonlinear optimizations, a series of works \cite{oleynikova2016continuous, Usenko2017ewok, ding2019efficient, zhou2019robust, zhou2020raptor} are proposed.
They consider smoothness, feasibility, and safety of the trajectory using various parameterization methods, including polynomial and B-spline.
Recently, we propose a single-quadrotor navigation system named EGO-Planner \cite{zhou2020ego}, which further reduces computation time using a more compact environment representation.
This is the work on which this paper is based.

\subsection{Topological Planning}
\label{sec:related_work_topological_planning}
Topological planning is used to escape local minima. 
Based on the homology equivalence relation in 2-D surfaces originated from complex analysis\cite{bhattacharya2010search}, R{\"o}smann et al.\cite{rosmann2017integrated} present a trajectory planning method in distinctive topologies using Voronoi and sampling-based front-ends and TEB (Timed-Elastic-Bands) local planner \cite{rosmann2012trajectory} as back-ends.
However, homology equivalence relation in 3-D is far more trivial.
To capture distinctive useful paths, Jaillet et al.\cite{jaillet2008path} construct visibility deformation roadmaps that encode richer and more relevant information than representative paths of the homotopy classes.
%Nevertheless, high computation burden restricts its application of trajectory generation.
Based on \cite{jaillet2008path}, Zhou et al.\cite{zhou2020robust} enable real-time topological planning by proposing an efficient topology equivalence checking.
We extend EGO-Planner to accelerate the front-end for topological planning further.

\subsection{Decentralized Drone Swarm}

Decentralized approaches have been proposed in, e.g., \cite{van2011reciprocalnbody, van2011reciprocalcollision, arul2020dcad, luis2019trajectory, park2020efficient}. 
Velocity obstacles is leveraged to guarantee collision-free trajectories for point robots \cite{van2011reciprocalnbody}, holonomic agents \cite{van2011reciprocalcollision}, and non-holonomic agents \cite{arul2020dcad}. 
Liu et al. \cite{liu2020mapper} propose a decentralized and asynchronous planning strategy for drones to avoid static/dynamic obstacles and inter-vehicle collisions. 
While those algorithms are validated through simulations without integrating sensing, mapping, and planning capabilities. 
Experimental results have been shown in \cite{luis2019trajectory, park2020efficient}. 
\cite{luis2019trajectory} enables multi-vehicle point-to-point transitions without external obstacles, and \cite{park2020efficient} relies on ad-hoc planning priority.
However, none of them achieve full autonomy in field environments.

\section{Implicit Topological Trajectory Generation of Gradient-based local planning}

In this section, we introduce our previous work on EGO-Planner\cite{zhou2020ego} at first, on which the proposed swarm system is based.
Then the proposed topological planning strategy is explained.

\subsection{An ESDF-Free Gradient-based local planner}
\label{sec:G_based_lp}

As a gradient-based local planner, EGO-Planner formulates trajectory generation as a non-linear optimization problem that trades off smoothness $J_s$, collision $J_c$, dynamically feasibility $J_d$, and terminal progress $J_t$.
The optimization problem, with decision variables from control points $\mathbf{Q}$ of a uniform B-spline $\mathbf{\Phi}$ used to parameterize the trajectory, is given by
\begin{equation}
\label{equ:cost_no_swarm}
\mathop{min}_{\mathbf{Q}} J_{\rm{EGO}} = \sum \lambda_r J_r,
\end{equation}
where $r=\{s,c,d,t\}$ and the subscripted $\lambda$ indicates the corresponding weights.
Terms of $J$ can be divided into two categories: \textit{minimum error} and \textit{soft barrier constraint}.
Minimum error terms $J_s$ and $J_t$, which minimize the total error between a linear transformation of decision variables $L(\mathbf{Q})$ and a desired value $\mathcal{D}$, are formulated as
\begin{equation}
J_r = \sum_{\mathbf{Q} \in \mathbf{\Phi}} \left\|L(\mathbf{Q}) - \mathcal{D}\right\|_n^n.
\end{equation}
Soft barrier constraint terms $J_c$ and $J_d$, which penalize decision variables exceeding a specific threshold $\mathcal{T}$, are generally expressed as
\begin{equation}
J_r = \sum_{\mathbf{Q} \in \mathbf{\Phi}} \left\{
\begin{array}{cl}
{\left\| \frac{L(\mathbf{Q})-(\mathcal{T}-\epsilon)}{S} \right\|}_n^n & L(\mathbf{Q}) > (\mathcal{T}-\epsilon) \\
0 & L(\mathbf{Q}) \leq (\mathcal{T}-\epsilon)
\end{array} \right.,
\end{equation}
where parameters $S$, $n$, and $\epsilon$ affect unilateral constraint approximation accuracy as described in \cite{rosmann2012trajectory}.
Transformation $L(\cdot)$ and parameters are chosen according to penalty types.
The exact form of $L(\cdot)$ is omitted here due to limited paper length and can be found in \cite{zhou2020ego}.

\begin{figure}[t]
	\centering
	\begin{subfigure}{0.48\linewidth}
		\includegraphics[width=1.0\linewidth]{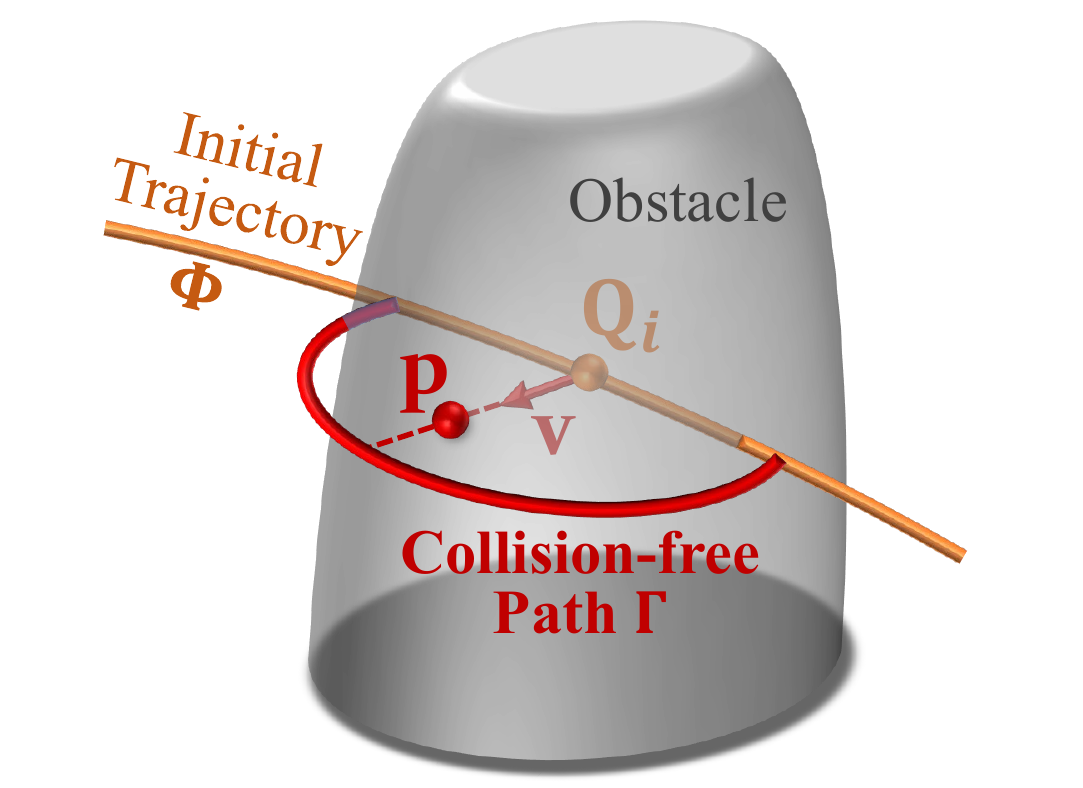}
		\captionsetup{font={small}}
		\caption{$\{\mathbf{p},\mathbf{v}\}$ Pair Generation}
		\label{pic:a_p_v_pair}	
	\end{subfigure}
	\begin{subfigure}{0.48\linewidth}
		\includegraphics[width=1.0\linewidth]{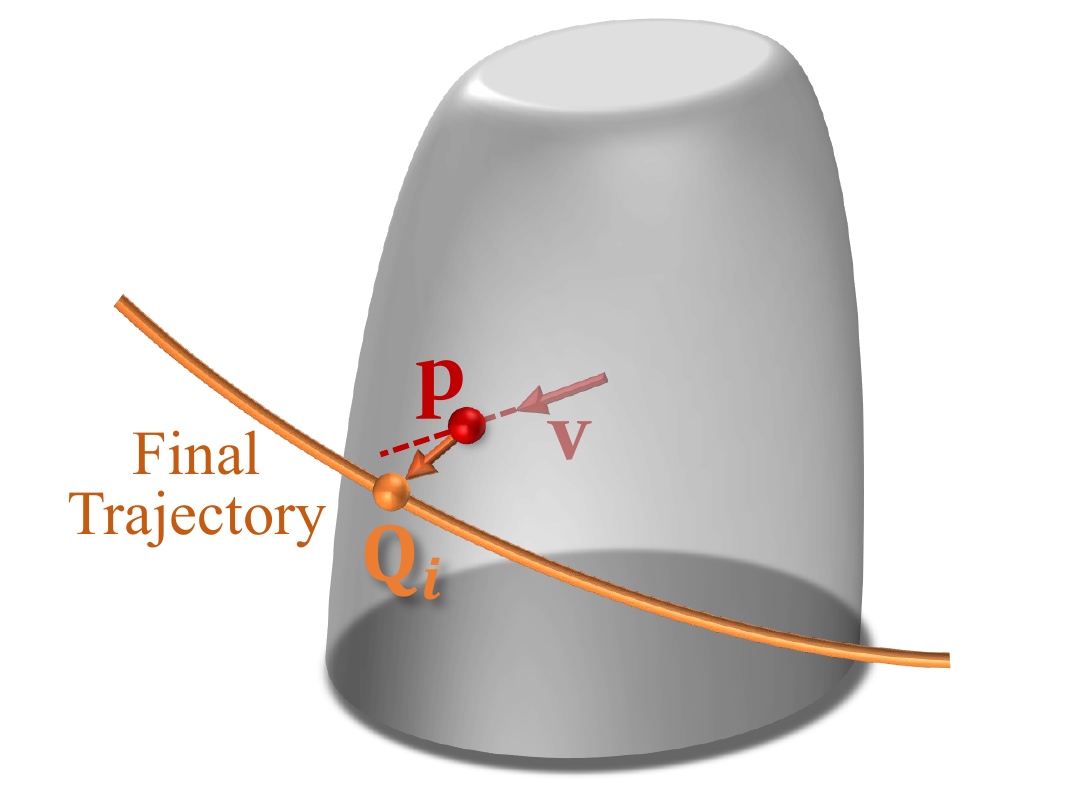}
		\captionsetup{font={small}}
		\caption{A Safe Trajectory Is Generated}
		\label{pic:final_traj}	
	\end{subfigure}
	\begin{subfigure}{0.48\linewidth}
		\includegraphics[width=1.0\linewidth]{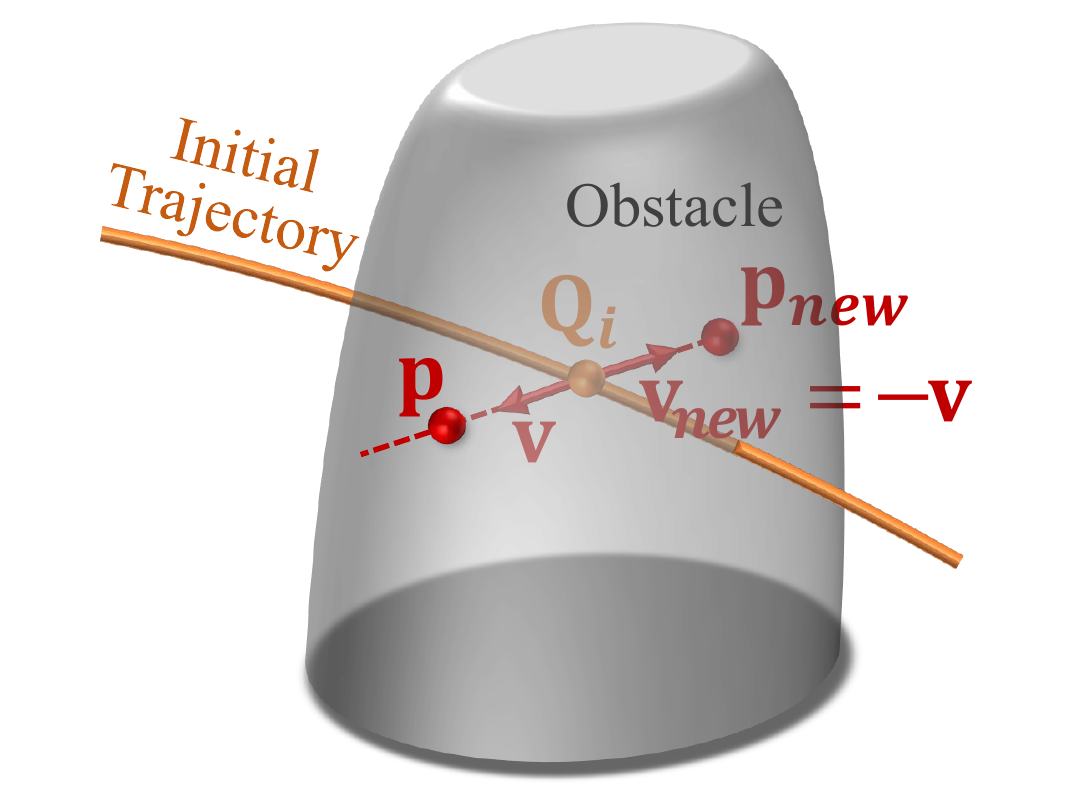}
		\captionsetup{font={small}}
		\caption{A New $\{\mathbf{p},\mathbf{v}\}$ Pair}
		\label{pic:multi_p_v}	
	\end{subfigure}
	\begin{subfigure}{0.48\linewidth}
		\includegraphics[width=1.0\linewidth]{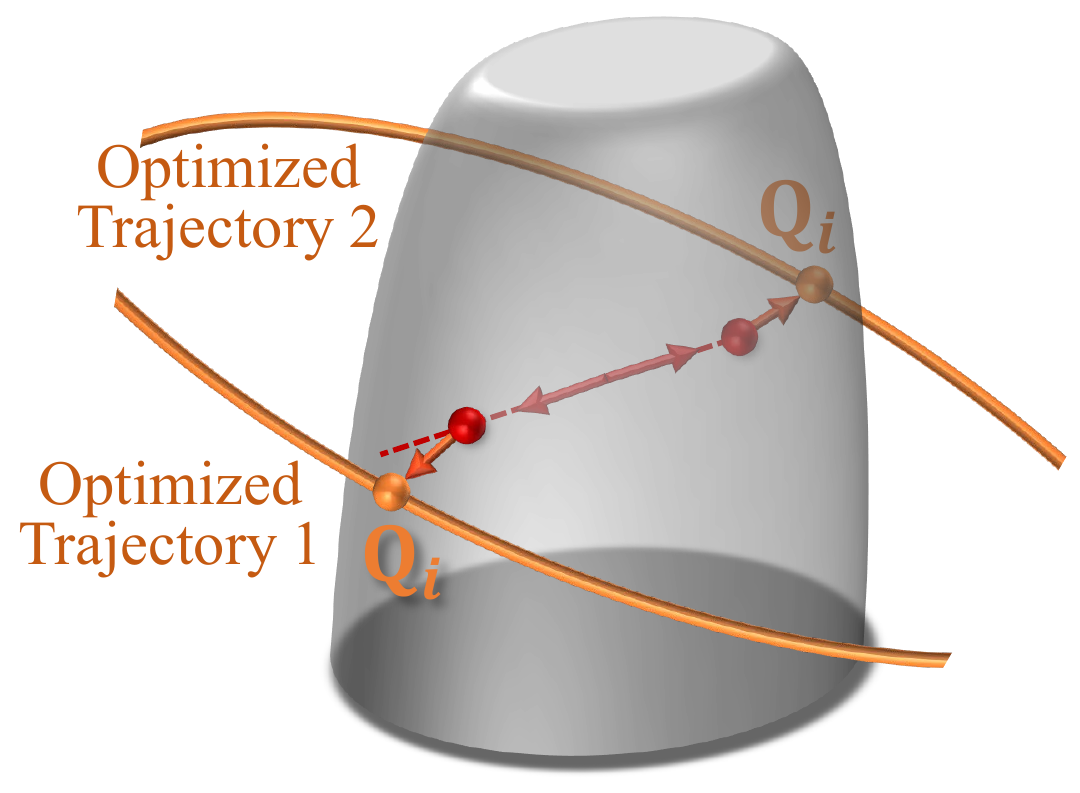}
		\captionsetup{font={small}}
		\caption{Trajectories at Opposite Sides}
		\label{pic:multi_traj}	
	\end{subfigure}
	\captionsetup{font={small}}
	\caption{ Illustration of EGO-Planner and topological trajectory generation. (a) A safe path bypassing a colliding trajectory is searched, then a $\{\mathbf{p},\mathbf{v}\}$ pair is generated. (b) The optimizer finds a safe trajectory that satisfies $d=(\mathbf{Q}_i - \mathbf{p}) \cdot \mathbf{v}$ grater than a constant value. (c) A new $\{\mathbf{p},\mathbf{v}\}$ pair with $\mathbf{v_{new}:=-\mathbf{v}}$ and $\mathbf{p_{new}}$ at the opposite side of the obstacle is generated. (d) A different trajectory satisfying the constraint of $\{\mathbf{p}_{new},\mathbf{v}_{new}\}$ pair is generated. }
	\vspace{-0.2cm}
\end{figure}

In EGO-Planner, we propose a novel obstacle distance estimation method from environment information possessed by each $\mathbf{Q}$ independently.
The information parameterized by several $\{\mathbf{p},\mathbf{v}\}$ pairs is highly abstracted from surrounding obstacles, where $\mathbf{p}$ denotes an anchor point at the obstacle surface and $\mathbf{v}$ represents a safe direction pointing from inside to outside of that obstacle, as illustrated in Fig.\ref{pic:a_p_v_pair}.
Then the obstacle distance $d_{ij}$ of $i^{th}$ control point $\mathbf{Q}_i$ to $j^{th}$ obstacle is defined as 
\begin{equation}
d_{ij}=(\mathbf{Q}_i - \mathbf{p}_{ij}) \cdot \mathbf{v}_{ij}.
\end{equation}

$\{\mathbf{p},\mathbf{v}\}$ pair generation and trajectory optimization procedure are shown in Fig.\ref{pic:a_p_v_pair} and \ref{pic:final_traj}. 
First, a naive initial trajectory $\mathbf{\Phi}$ is given, regardless of collision.
Then a safe path $\mathbf{\Gamma}$ connecting two ends of the colliding segment of $\mathbf{\Phi}$ is searched.
After that, vector $\mathbf{v}$ is generated from $\mathbf{\Phi}$ to $\mathbf{\Gamma}$ and $\mathbf{p}$ is defined at the obstacle surface.
With generated $\{\mathbf{p},\mathbf{v}\}$ pairs, the planner maximizes $d_{ij}$ and returns an optimized trajectory.
Here we only give a simplified description of the basic idea of EGO-Planner due to the limited article length.
Detailed explanations can be found in \cite{zhou2020ego}.

\begin{figure}[t]
	\centering
	\includegraphics[width=0.9\linewidth]{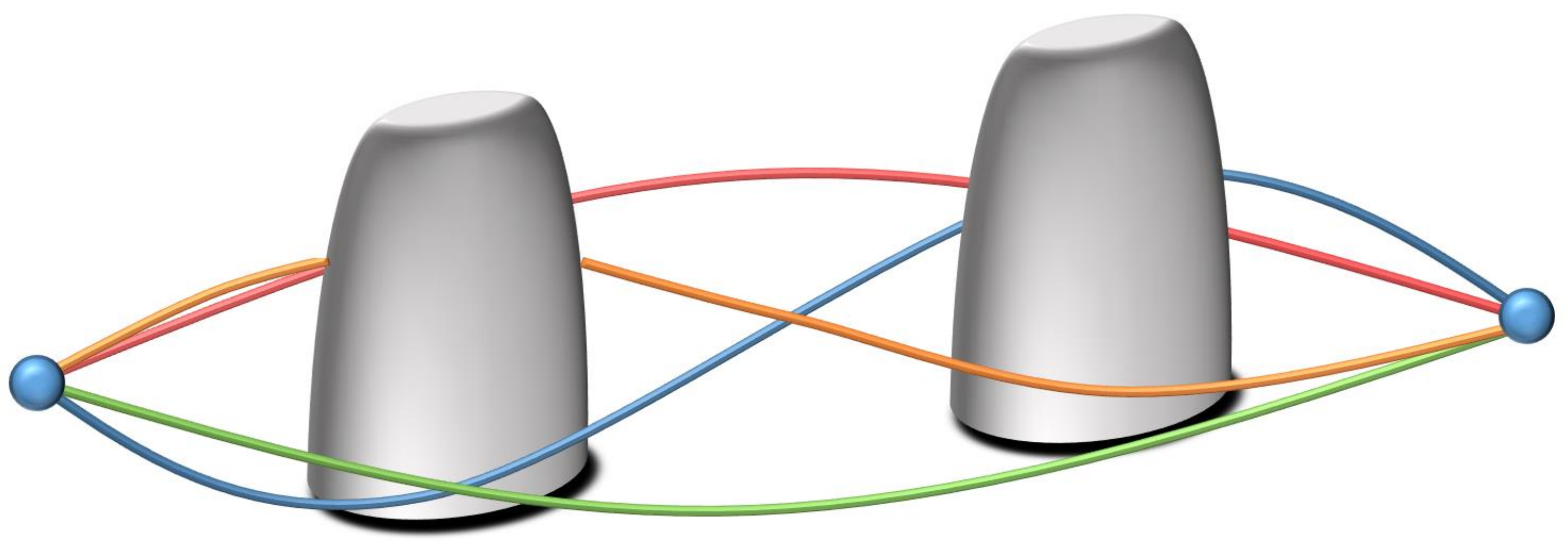}
	\captionsetup{font={small}}
	\caption{ Four trajectories belonging to the same 3-D homotopy stay in different local minima, which we aim to seek. }
	\label{pic:topo_relation}
	\vspace{-1.3cm}
\end{figure}

\subsection{Implicit Topological Trajectory Generation}
Analysis in \cite{jaillet2008path, zhou2020robust} reveals that the widely used homotopy concept is insufficient to capture candidate trajectories in 3-D cases, as shown in Fig.\ref{pic:topo_relation}.
Therefore, Jaillet et al.\cite{jaillet2008path} propose a more useful relation in 3-D space named visibility deformation (VD), and Zhou et al. \cite{zhou2020robust} further extract a subset of VD called uniform visibility deformation (UVD), which enables real-time operation.
However, we still use the term \textit{topological planning} in this paper as previous works do without ambiguity.
Trajectories satisfying UVD are considered homeomorphic.
UVD defined in \cite{zhou2020robust} is

\newtheorem{definition}{\textbf{Definition}}
\begin{definition}
	\label{def:UVD}
	Two trajectories $\tau_1(s)$, $\tau_2(s)$, parameterized by $s \in \left[0,1\right]$ and satisfying $\tau_1(0) = \tau_2(0)$, $\tau_1(1) = \tau_2(1)$, belong to the same UVD class, if for all $s$, line $\tau_1(s) \tau_2(s)$ is collision-free.
\end{definition}

Traditional topological planning methods \cite{jaillet2008path, rosmann2012trajectory, rosmann2017integrated, zhou2020robust}, which consist of topologically distinct path search and back-end optimization, focus mainly on finding multiple initial paths in distinct homotopy.
Unlike those methods, the proposed method constructs distance fields in different directions by inverting $\mathbf{v}$ to $\mathbf{v}_{new} := -\mathbf{v}$.
Then a search process determines a new anchor point $\mathbf{p}_{new}$ at the obstacle surface alone $\mathbf{v}_{new}$, as depicted in Fig.\ref{pic:multi_p_v}.
They constitute a new pair $\{\mathbf{p}_{new}, \mathbf{v}_{new}\}$, which leads to a different local minima.
Note that no explicit path search is adopted, but any pair of paths through $\mathbf{p}$ and $\mathbf{p}_{new}$ respectively violates Def.\ref{def:UVD} at these two points naturally.
Subsequently, distinctive trajectories are optimized in parallel in different threads, as illustrated Fig.\ref{pic:multi_traj}.
The trajectory with lowest cost is executed.

\section{Drone Swarm Navigation}

\subsection{Reciprocal Collision Avoidance}
\label{sec:Reciprocal_Collision_Avoidance}
\begin{figure}[t]
	\centering
	\includegraphics[width=1.0\linewidth]{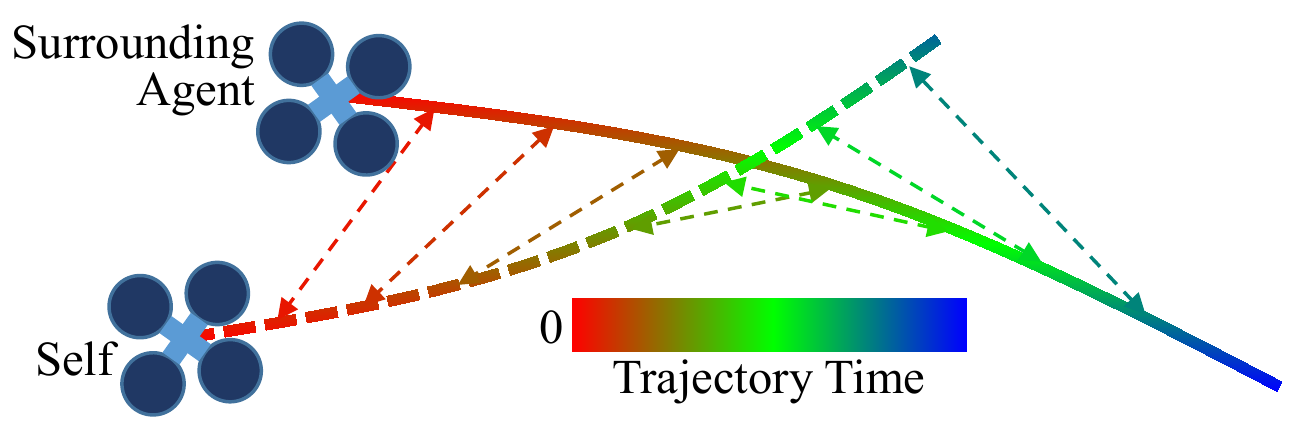}
	\captionsetup{font={small}}
	\caption{ Drone \textit{Self} generates a trajectory by comparing the distance with a received trajectory from a surrounding drone of the same trajectory time. }
	\label{pic:distance_test}	
	\vspace{-1.0cm}
\end{figure}

%Reciprocal collision-free is defined as 
Let $x_k(t) \in \mathcal{X} \subset \mathbb{R}^3$ be the position state of the agent $k$ among $K$ agents at time $t$. 
$\mathcal{X}_k^{free}(t) \subset \mathcal{X}$ is the free region in the state space of agent $k$ considering the existence of other agents.
Thus, $\mathcal{X}_k^{free}(t) := \mathcal{X} \backslash \{ i \in \mathbb{Z} \backslash k, i \leq K | x_i(t)\}$ and the valid trajectory $\mathbf{\Phi}_k$ satisfies for any $t$ within the domain of $\mathbf{\Phi}_k$, $\mathbf{\Phi}_k(t) \in \mathcal{X}_k^{free}(t)$, as depicted in Fig.\ref{pic:distance_test}.
Unlike \cite{liu2017search}, here we ignore obstacles and dynamic constraints which have already get tackled in Sec.\ref{sec:G_based_lp}.

Similar to the penalty on obstacle collision and dynamical infeasibility, we formulate the penalty function $J_{w,k}$ of swarm collision avoidance for agent $k$ as \textit{soft barrier constraint} 
\begin{equation}
\label{equ:swarm_collision}
\begin{array}{l}
J_{w,k} = \sum_i \int_{t=t_s}^{t_e} \left\{ 
\begin{array}{cl}
d_{k,i}(t)^2 \\
0 
\end{array} \right. \! \! 
dt, 
\begin{array}{cl}
d_{k,i}(t) < 0 \\
\! d_{k,i}(t) \geq 0
\end{array} \\
d_{k,i}(t) = \left\| \mathbf{E}^{1/2}\left[\mathbf{\Phi}_k(t) - \mathbf{\Phi}_i(t)\right]\right\| - (\mathcal{C}+\epsilon)
\end{array},
\end{equation}
where $i \in \mathbb{Z} \backslash k, i \leq K$, $t_s$ and $t_e$ are global start and end time within the time span of trajectory $\mathbf{\Phi}_k(t)$.
$\mathcal{C}$ is user-defined agent clearance.
$\mathbf{E}:=\rm{diag}(1,1,1/c), c>1$ transforms Euclidean distance into \textit{ellipsoidal distance} with shorter principal axes at z-axis to relief downwash risk.
Adding the weighted $J_{w,k}$ to Equ. (\ref{equ:cost_no_swarm}) yields the total optimization problem for each agent
\begin{equation}
\mathop{min}_{\mathbf{Q}} J = J_{\rm{EGO}} + \lambda_w J_{w}.
\end{equation} 

Any trajectory parameterization method containing a mapping from decision variables to points on the trajectory is applicable to Equ. (\ref{equ:swarm_collision}).
This paper parameterizes trajectory using a $p_b$-degree uniform B-spline, which gives the position evaluation a matrix representation\cite{zhou2019robust} 
\begin{equation}
\begin{array}{l}
\mathbf{\Phi}(t) = \mathbf{s}(t)^{\top}\mathbf{M}_{p_b+1}\mathbf{q}_m \\
\mathbf{s}(t) = \left[1 \quad s(t) \quad s^{2}(t) \quad \cdots \quad s^{p_b}(t) \right]^{\top}\\
\mathbf{q}_m = \left[\mathbf{Q}_{m-p_b} \ \mathbf{Q}_{m-p_b+1} \ \mathbf{Q}_{m-p_b+2} \ \cdots \ \mathbf{Q}_{m}\right]^{\top}
\end{array},
\end{equation}
where $\mathbf{M}_{p_b+1}$ is a constant matrix determined by $p_b$, $s(t)=(t-t_m)/\triangle t$ when $t$ belongs to the knot span $(t_m, t_{m+1}]$.

\subsection{Localization Drift Compensation}
\label{sec:compensate_localization_drafting}
As a result of individual localization in unknown environments (no reliable and high-frequency loop-closure), drift accumulates during the flight.
Xu et al. \cite{xu2020decentralized} propose a state estimation method for aerial swarms with extra UWB distance measurements and achieve accurate collaborative localization.
However, we focus more on traversing obstacle-rich environments and have to reserve computation and communication resources for other applications. 
Therefore, inspired by \cite{xu2020decentralized}, a simplified and lightweight relative drift estimation method is proposed by comparing the predicted position evaluated from received agents' trajectories and the measured positions from depth images of witnessed agents.
This strategy works when the trajectory tracking error is negligible and at least one of any two agents that might collide see the other.
Therefore, we use the controller from \cite{MelKum1105} for precise tracking and use wide-angle cameras to reduce the possibility of losing agents.

The drift elimination procedure is as follows.
After evaluating the current position $\mathbf{\Phi}_i(t_{now})$ of agent $i$, a spherical trust region $\mathcal{S} \subset \mathbb{R}^3 $ centered at $\mathbf{\Phi}_i(t_{now})$ with radius $\mathcal{R}$ is determined, where $\mathcal{R}$ is an empirical parameter indicating the upper bound of typical drift estimated from experiments. 
$\mathcal{S}$ is then mapped to the currently captured depth image, which is the region $\mathcal{S'} \subset \mathbb{R}^2 $ satisfying 
\begin{equation}
z\left[\mathbf{s'}^{\top} \ 1\right]^{\top} \!\! = \mathbf{K}\mathbf{T}^{\rm{c}}_{\rm{w}} \left[\mathbf{s}^{\top} \ 1 \right]^{\top},
\end{equation}
where $\mathbf{s'} \in \mathcal{S'}$, $\mathbf{s} \in \mathcal{S}$, $\mathbf{K}$ and $\mathbf{T}^{\rm{c}}_{\rm{w}}$ are camera intrinsic and extrinsic matrices, $z$ is the deviation of ${s}$ along the main optical axis from the optical center.
$\mathcal{S'}$ is an elliptical conic section which requires complex calculation to acquire.
Therefore, we adopt an approximate axis-aligned ellipse $\mathcal{\bar{S}'}$ instead of the exact $\mathcal{S'}$.
There is no necessity to precisely define the trust region since it is just an empirical region.

Then we project each point within $\mathcal{\bar{S}'}$ into the world frame and collect points belonging to $\mathcal{S}$, which results in a point cluster $\mathcal{P} \subset \mathcal{S}$.
Then the position of agent observation $\mathbf{P}$ is regarded as the center (first raw moment) of $\mathcal{P}$, which is
\begin{equation}
	\label{equ:moment}
	\mathbf{P} = \mu'_1(\mathcal{P}).
\end{equation}
Equ. (\ref{equ:moment}) holds if $\mathcal{P}$ contains only the corresponding agent's observation without any unrelated object, which is not guaranteed.
However, since each agent plans trajectories with a clearance to nearby objects, Equ. (\ref{equ:moment}) holds for most of the time.
Additional criteria are added to improve the robustness for agent detection, such as the number of pixels, the second central moment of $\mathcal{P}$, the deviation of the current measurement to previous ones, etc.
Stricter criterion increases false-negative rate, but it is harmless since localization drift alters slowly.
Finally, the error between $\mathbf{\Phi}_i(t_{now})$ and $\mathbf{P}$ is fed to a filter, from where the estimated drift is then acquired. 

\subsection{Agent Removal from Depth Images}
\begin{figure}[t]
	\centering
	\includegraphics[width=1\linewidth]{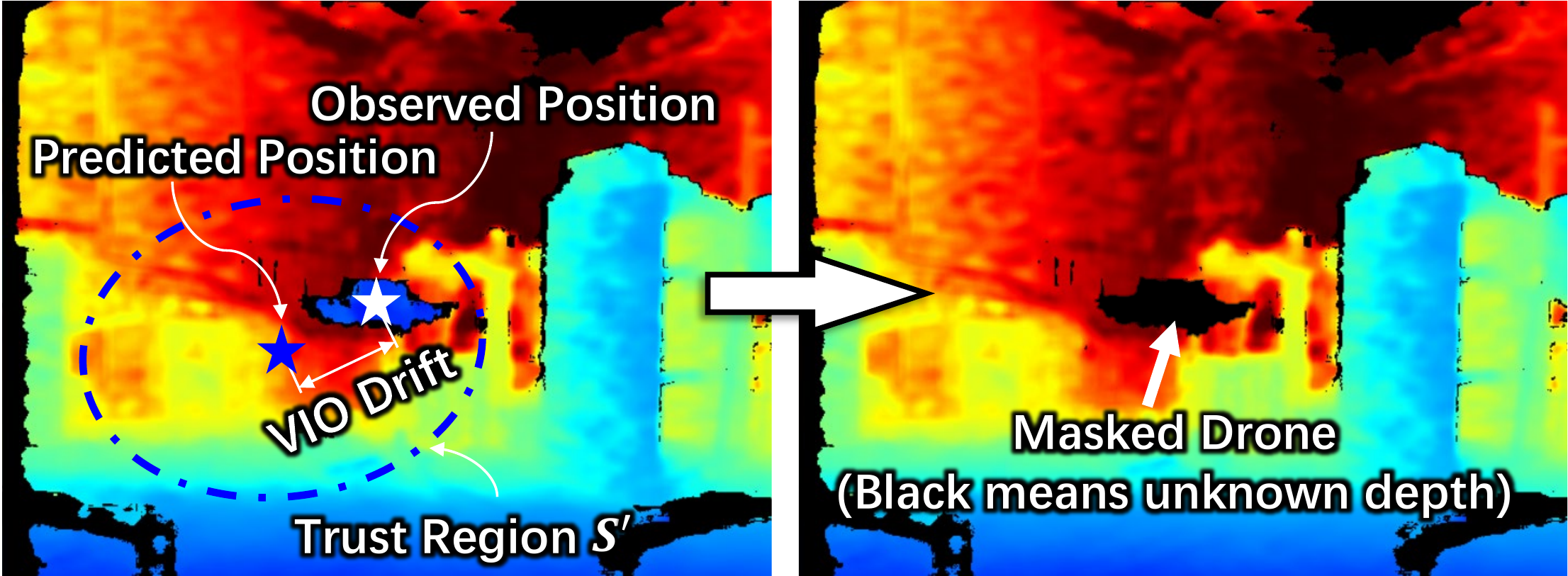}
	\captionsetup{font={small}}
	\caption{ \textbf{Left:} VIO (the localization method we used) Drift is estimated between a predicted position evaluated from the received trajectory and an observed position in the depth image within the trust region. \textbf{Right:} The observed agent is masked out to avoid affecting mapping. }
	\label{pic:mask}	
	\vspace{-0.3cm}
\end{figure}

We use occupancy grid map to store static obstacles, and depth images for map fusion.
The moving agents are taken care of in Sec.\ref{sec:Reciprocal_Collision_Avoidance}. 
Therefore, recording moving agents and treating them as obstacles in the map building are not necessary, even harmful.
To eliminate the influence of moving objects, we mask and remove the pixels of agents detected in Sec.\ref{sec:compensate_localization_drafting} from the depth images, as illustrated in Fig.\ref{pic:mask}.
Except that, moving objects covering the majority of the view are interference to VIO. 
Therefore, agents on gray-scale images are removed as well using the same mask for the corresponding depth images.
The criterion for agent detection used here is less strict since false positives are more harmful than false negatives.

\begin{figure}[t]
	\centering
	\includegraphics[width=1.0\linewidth]{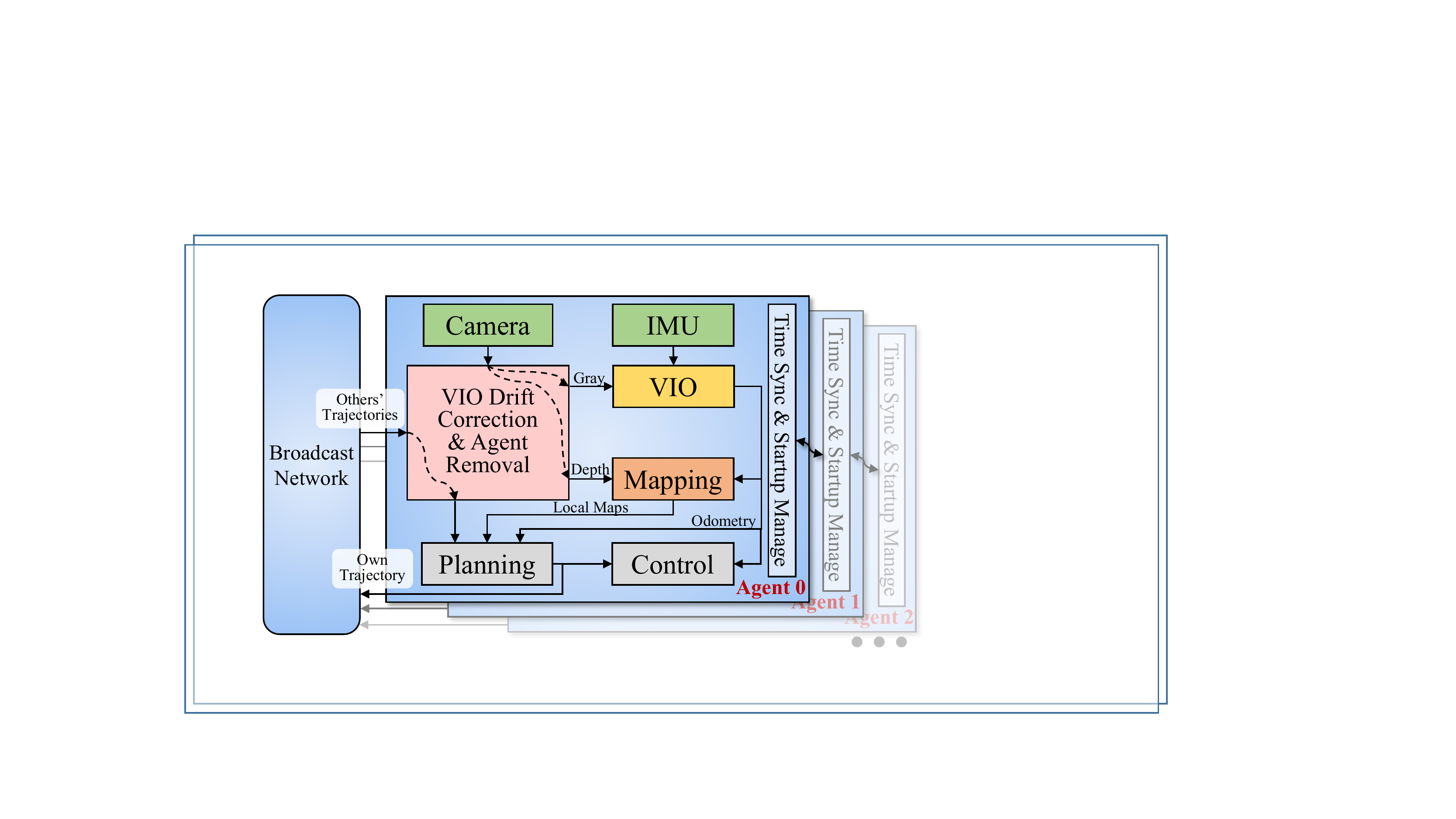}
	\captionsetup{font={small}}
	\caption{ System Architecture }
	\label{pic:system_architecture}	
	\vspace{-1.5cm}
\end{figure}

\section{System Architecture}
The system architecture is depicted in Fig.\ref{pic:system_architecture}, which contains the detailed architecture for a single agent and the multi-agent communication system.

\subsection{Navigation System of A Single Agent}
The single agent system, including hardware and software setups is based on our previous work EGO-Planner \cite{zhou2020ego}, with an extra module that compensates VIO drift and removes witnessed agents on images.
%We use VINS-fusion \cite{qin2018vins} for localization, occupancy grid map for mapping, and the controller from \cite{MelKum1105}.
%The hardware is composed of Intel Realsense D435\footnote{https://www.intelrealsense.com/depth-camera-d435} camera to acquire both depth and gray-scale images, DJI N3\footnote{https://www.dji.com/n3} as the low level flight controller with IMU output, and DJI minifold-2C\footnote{https://www.dji.com/manifold-2} computer for onboard calculations.
%All the components are integrated into a self-assembled 250mm wheelbase quadrotor.
For trajectory generation in unknown environments, the local planner is used.
Planning is activated when the current trajectory collides with newly discovered obstacles, or the agent is getting close to the end of the current trajectory.

\subsection{Communication Framework}
Two networks connect the system, i.e., a broadcast network sharing trajectories, and a chain network to synchronize timestamps and manage the sequential startup.

\begin{figure}[t]
	\centering
	\includegraphics[width=1.0\linewidth]{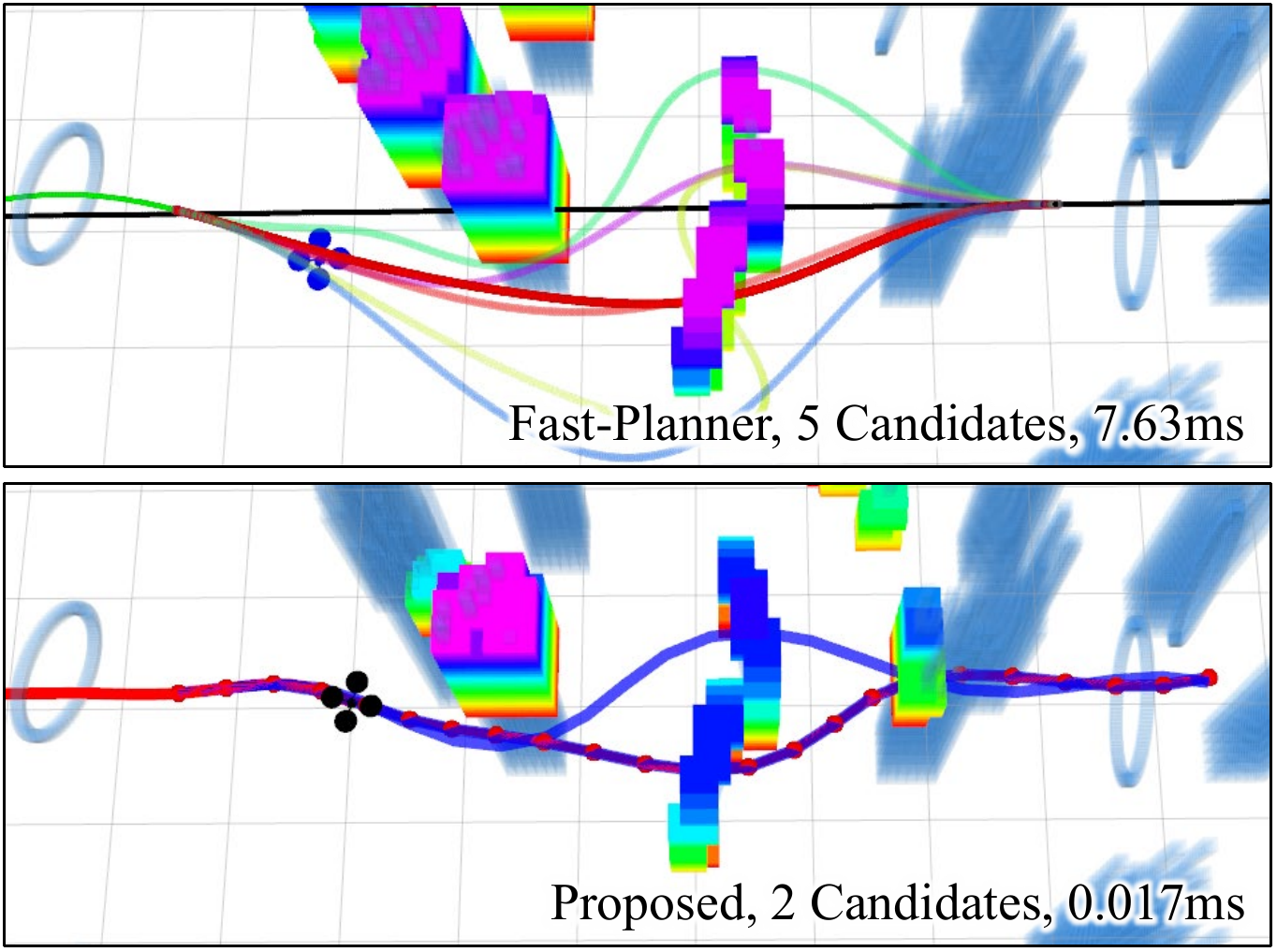}
	\captionsetup{font={small}}
	\caption{ A comparison of front-end topological path search. The proposed planner finds fewer candidate topologically distinct trajectories but consumes less computation compared to Fast-Planner. Note that the time is only for front-end local minima discovery, while displayed trajectories are optimized for better visualization.}
	\label{pic:topo_comp}	
	\vspace{-1.0cm}
\end{figure}

\subsubsection{Broadcast Network}
\label{sec:Broadcast_Network}
Once an agent generates a new collision-free trajectory, it is broadcast immediately to all agents.
Other agents then receive and store this trajectory used to generate safe trajectories for themselves when necessary.
This closed-loop strategy works properly in ideal situations where the connection is stable and the latency is negligible.
However, this is not guaranteed in practice.
Therefore, we propose two methods to reduce the possibility of collision.

First, one trajectory is broadcast at a given frequency under network capacity.
This does not cause computation burden since a typical trajectory containing 3-D waypoints and other parameters are less than 0.5KB in size.
By contrast, modern wireless networks such as Bluetooth\footnote{https://www.bluetooth.com/} can reach a speed above 1Mbps.
Second, each agent checks collision as soon as a trajectory is received from broadcast network, and if potential collisions are detected, a new collision-free trajectory is generated.
This strategy can tackle when multiple agents generate trajectories at a close time without receiving others' trajectories due to latency or packet losses.

In addition, the growth of computational complexity with the increase of agent number is considered.
Before planning, each agent compares its current position with received surrounding agents' trajectories, where any trajectory outside the planning range will be ignored.

\subsubsection{Chain Network}
A connection-based stable chain network is used for timestamp synchronization and system startup management.
At system startup, agents generate trajectories in a predefined order.
Each agent generates its initial trajectory after receiving trajectories from agents with higher priority through the chain network.
This strategy avoids chaos caused by simultaneous trajectory generation during system startup since agents have no information of other trajectories at that time.

\begin{table}[ptb]
	%	\vspace{0.5cm}
	\centering
	\caption{Comparisons in Obstacle-free Space }  
	\begin{tabular}{c|cccc}
		\toprule
		Planner & $d_{\rm{fly}}$(m) & $t_{\rm{fly}}$(s) & Collision & $t_{\rm{cal}}{}^*$ \\
		\midrule
		DMPC & 20.6 & \bf 10.8 & \bf 0 & 0.668 (s) \\
		RBP & 22.9 & 23.6 & \bf  0 & 1.18 (s)  \\
		ORCA & 21.3 & 28.5 & 0.375 & \bf 0.0225 (ms) \\
		Proposed & \bf 20.3 & 14.3 & \bf  0 & 0.514 (ms)  \\
		\bottomrule
	\end{tabular}
	\label{tab:plan_cmp}
		\vspace{-1.3cm}
\end{table} 

\section{Benchmark}
Benchmark comparisons are presented in simulation using an i7-9700KF CPU. For parameter settings, the planning horizon is set to 7.5m. $\lambda_s=1.0,~\lambda_c=\lambda_w=\lambda_t=0.5,~\lambda_d=0.1$. Map resolution is 0.1m. Re-planning is triggered every second or a collision is predicted. This setting is implemented in both simulation and real-world experiments.

\subsection{Topological Planning}
We compare the topological planning performance of the proposed EGO-Swarm with Fast-Planner\cite{zhou2020robust} in terms of candidate trajectory number and computation time for front-end topological path search.
As is shown in Fig.\ref{pic:topo_comp}, EGO-Swarm finds fewer candidate trajectories, which means a lower probability of finding the global optimal, but is faster than \cite{zhou2020robust} by two orders of magnitude.
Since Fast-Planner finds topologically distinct paths by PRM\cite{kavraki1996probabilistic} graph search, path shortening, and path pruning, which are time-consuming but with a higher degree of freedom, compared to the proposed implicit topological path search method.

%\begin{table}[ptb]
%%\vspace{0.5cm}
%\centering
%\caption{Comparison of Topological Path Search } 
%\begin{tabular}{c|ccc}
%	\toprule
%	Planner & Candidates & $\int (jerk)^2 dt$ & $t_{cal}$(ms) \\
%	\midrule
%	Fast-Planner & \bf5 & \bf 37.5 & 8.63 \\
%	Proposed & 2 & 62.3 & \bf  0.017 \\
%	\bottomrule
%\end{tabular}
%\label{tab:topo_cmp}
%%\vspace{-1.0cm}
%\end{table}

\subsection{Swarm Planning}
\label{sec:benchmark}

\begin{figure*}[t]
	\centering
	\includegraphics[width=1.0\linewidth]{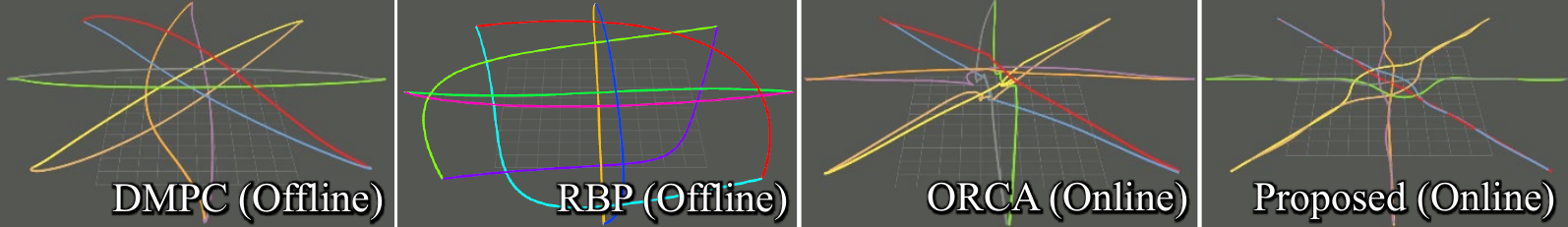}
	\captionsetup{font={small}}
	\caption{ Planned trajectories by various swarm planners in obstacle-free scenarios.  }
	\label{pic:multi_comp}	
	\vspace{-0.2cm}
\end{figure*}

\subsubsection{In Empty Space}
We compare the proposed method with DMPC\cite{luis2019trajectory}, ORCA\cite{van2011reciprocalnbody}, and RBP\cite{park2020efficient}, in terms of flight distance($d_{\rm{fly}}$), flight time($t_{\rm{fly}}$), collision times per agent, and computation time($t_{\rm{cal}}$).
The default parameters, except for maximum velocity and acceleration, are used for each compared method.
As depicted in Fig.\ref{pic:multi_comp}, eight agents perform swap transitions on a circle.
%, so that collision avoidance is necessary in such cases.
The results in Tab.\ref{tab:plan_cmp} are the average of all agents.
%Among four methods, DMPC and RBP generate trajectories offline, while the proposed EGO-Swarm and ORCA perform planning during the flight.
$t_{\rm{cal}}$ is marked by '*', since the computation time we record for offline methods, DMPC and RBP, is the total time to plan all agents' whole trajectory, while for ORCA and EGO-Swarm, it is the local replanning time for each agent.

Tab.\ref{tab:plan_cmp} and Fig.\ref{pic:multi_comp} state that RBP tends to generate safe but conservative trajectories, since the construction of convex \textit{relative safe flight corridor}\cite{park2020efficient} significantly compresses the solution space.
%The overall quality of the trajectory generated by DMPC in this empty space is best.
%However, computing time restricts its application in real-time systems.
DMPC is designed for distributed deployment.
Nevertheless, it requires accurate and high-frequency pose communication, which can not be guaranteed in real-word applications.
Efficient rules make ORCA update fast.
However, using speed as control commands makes it incompatible with third-order systems, such as quadrotors.
The risk of collision also limits its application.
%By contrast, though the proposed EGO-Swarm system can't surpass all methods on every metric, but it has no bottleneck.
By contrast, the proposed method generates the shortest collision-free, non-conservative trajectories with fast computation.
Therefore, it enables real-time applications for quadrotors.

\subsubsection{In Obstacle-rich Environments}
\label{sec:In_Obstacle-rich_Environments}
We simulate ten drones flying from one side of the map to the other with a speed limit of $2m/s$ and the quadrotor radius of 0.2 meters.
Fig.\ref{pic:sim} is a simulation snapshot of 0.42 obstacles/$m^2$.
Each agent senses the environment independently, and the constructed local maps are displayed in different colors.
The results are summarized in Tab.\ref{tab:pfms_vs_obs}, where $d_{\rm{fly}}$ is the average flight distance, $d_{\rm{safe}}$ is the closest distance to obstacles during the flight tests.
An inverse point-to-point transition is designed to make reciprocal collision avoidance inevitable around the map center.
In this scenario, each agent belonging to the swarm manages to plan smooth and safe trajectories.

\begin{table}[ptb]
	%	\vspace{0.5cm}
	\centering
	\caption{ Performance of EGO-Swarm in Simulation of Various Obstacle Densities }  
	\begin{tabular}{c|cccc}
		\toprule
		Density(obs/$m^2$) & $d_{\rm{fly}}$(m) & vel.(m/s) & $d_{\rm{safe}}$(m) & $t_{\rm{cal}}$(ms) \\
%		\makecell[c]{Density \\ ($obs/m^2$)} & $d_{fly}$(m) & vel.(m/s) & $d_{obs}$(m) & $t_{cal}$(ms) \\
		\midrule
		0.42 & 49.4 & 1.61 & 0.208 & 5.00 \\
		0.28 & 45.4 & 1.59 & 0.22 & 2.87 \\
		0.14 & 43.7 & 1.57 & 0.266 & 1.21 \\
		0 & 42.2 & 1.55 & / & 0.49 \\
		\bottomrule
	\end{tabular}
	\label{tab:pfms_vs_obs}
		\vspace{-0.2cm}
\end{table} 

\subsubsection{Scalability Analysis}
\begin{figure}[t]
	\centering
	\includegraphics[width=1.0\linewidth]{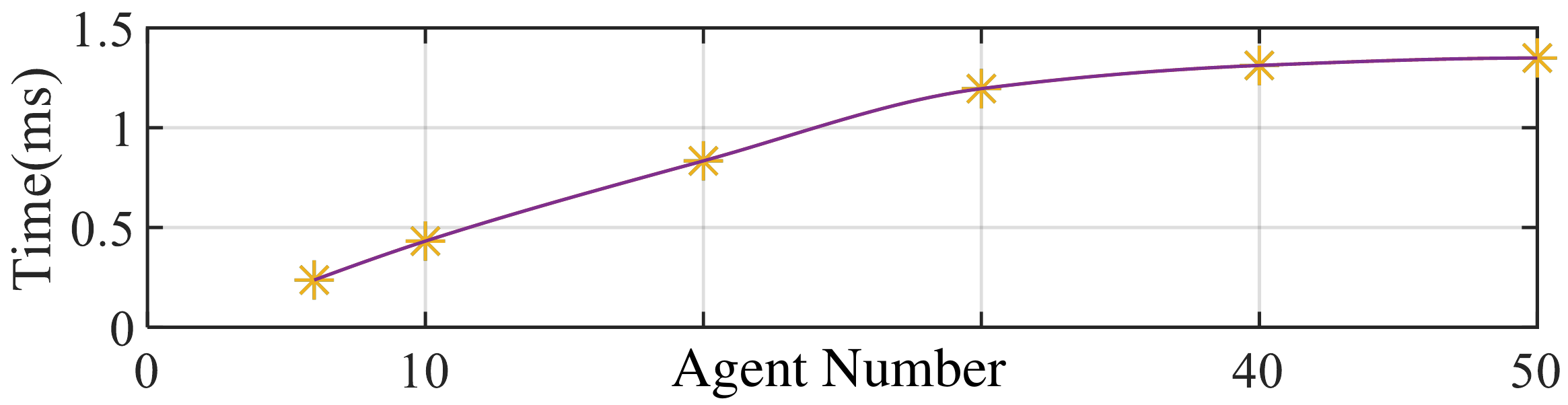}
	\captionsetup{font={small}}
	\caption{ The relationship between the computation time and the number of Agents in empty space. }
	\label{pic:computation_time}	
	\vspace{-1.4cm}
\end{figure}

We evaluate the computation performance in a scenario where agents arranged in a straight line fly to random target points 50 meters away. 
As is depicted in Fig.\ref{pic:computation_time}, due to the on-demand collision check strategy in Sec.\ref{sec:Broadcast_Network}, time complexity gradually flattens out with the increase in the number of agents.

%\vspace{0.5cm}

\section{Real-world Experiments}

\subsubsection{Indoor}
\begin{figure}[t]
	\centering
	\includegraphics[width=1\linewidth]{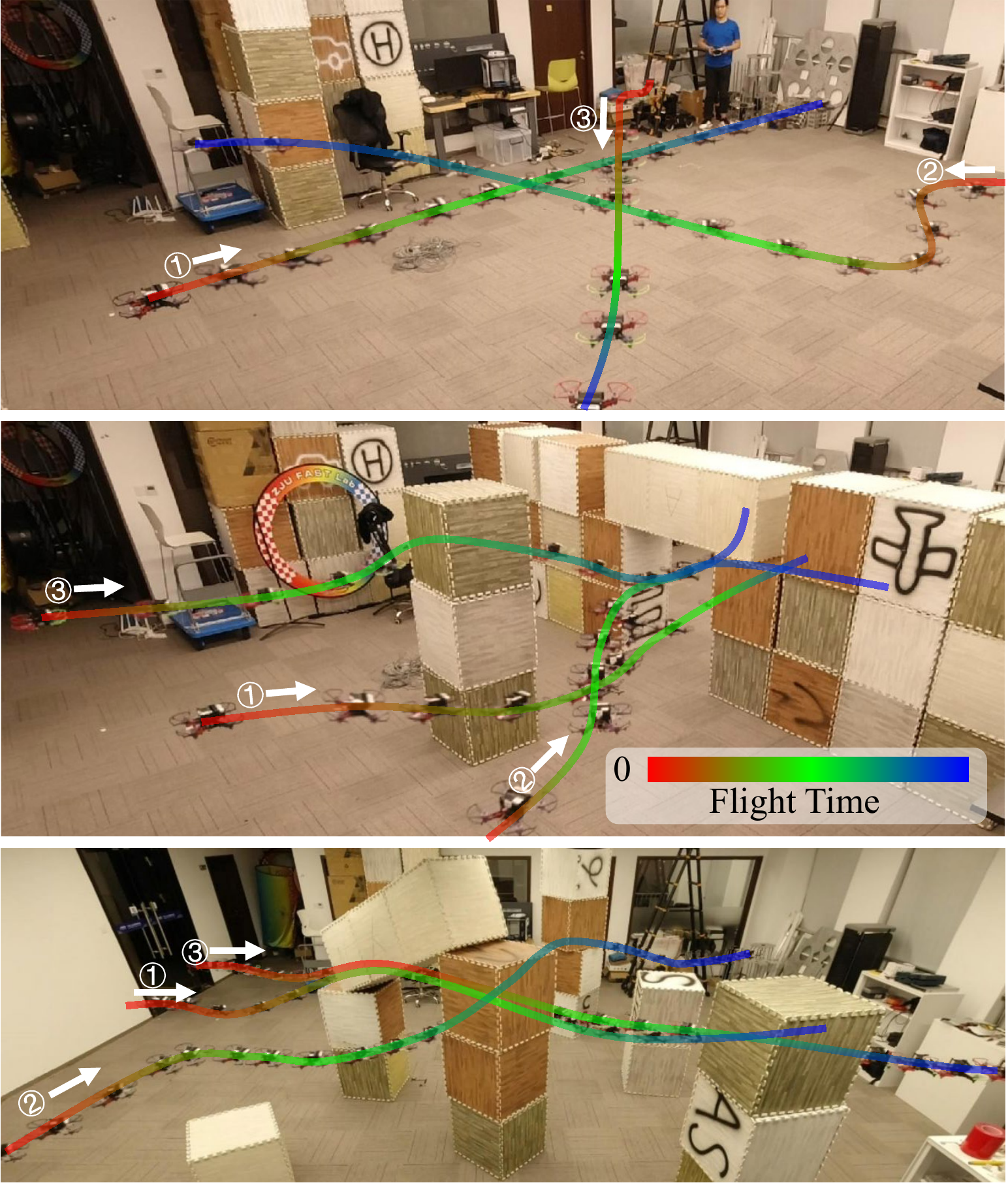}
	\captionsetup{font={small}}
	\caption{ Indoor Experiments }
	\label{pic:composed_image}
	\vspace{-1.0cm}	
\end{figure}

Indoor experiments are presented at a speed limit of 1.5m/s, as in Fig.\ref{pic:composed_image}.
%The color of the path indicates the approximate flight time of each quadrotor.
The top one shows three quadrotors perform a circle swap with reciprocal collision avoidance.
In the middle one, quadrotors manage to pass through a narrow door one after another.
In the bottom figure, the environment is more cluttered.
Three quadrotors manage to navigate across this environment.

\subsubsection{Outdoor}
In a forest where trees are spaced about 2 meters apart as shown in Fig.\ref{pic:forest}, three quadrotors start in the forest together and manage to reach the target position outside the forest.
The velocity limit is set to 1.5m/s.
To further emphasize reciprocal avoidance, we reverse the order of the goal position relative to the start position, making reciprocal avoidance unavoidable, like what we do in Sec.\ref{sec:In_Obstacle-rich_Environments}.
For more information about the experiments, please refer to the codes and watch our attached video on github\footnote{https://github.com/ZJU-FAST-Lab/ego-planner-swarm}.

\section{Conclusion}

In this paper, a systematic solution for multi-robot navigation in unknown cluttered environments using only onboard resources is proposed.
Benchmark comparisons demonstrate its short computation time and high trajectory quality.
Real-world experiments validate its robustness and efficiency.

%In the future, we will be committed to achieving precise co-localization and real-world dynamic obstacle avoidance.
%The main challenge lies in imprecise observations of other drones and unreliable moving object detection and prediction. 
%Finally, we hope to offer the robotics community a complete autonomous navigation solution for quadrotor swarms in complex environments.
%	

\newlength{\bibitemsep}\setlength{\bibitemsep}{.033\baselineskip}
\newlength{\bibparskip}\setlength{\bibparskip}{0pt}
\let\oldthebibliography\thebibliography
\renewcommand\thebibliography[1]{%
  \oldthebibliography{#1}%
  \setlength{\parskip}{\bibitemsep}%
  \setlength{\itemsep}{\bibparskip}%
}
\bibliography{icra2021Xin} 
\end{document}